\documentclass[10pt,twocolumn,letterpaper]{article}

\usepackage[pagenumbers]{cvpr} %

\usepackage[dvipsnames]{xcolor}

\definecolor{cvprblue}{rgb}{0.21,0.49,0.74}
\usepackage[pagebackref,breaklinks,colorlinks,citecolor=cvprblue]{hyperref}

\usepackage{mathabx}

\title{Still-Moving: Customized Video Generation without Customized Video Data \vspace{-0.5cm}}

\author{
\begin{tabular}{ccccc}
    Hila Chefer$^{\dagger \ 1 \ 2}$ &
    Shiran Zada$^{*\ 1}$ &
    Roni Paiss$^{*\ 1}$ &
    Ariel Ephrat$^{*\ 1}$ &
    Omer Tov$^{*\ 1}$\vspace{-0.3cm} \\
    \\
    Michael Rubinstein$^{\ 1}$ &
    Lior Wolf$^{\ 2}$ &
    Tali Dekel$^{\ 1 \ 3}$ &
    Tomer Michaeli$^{\ 1 \ 4}$ &
    Inbar Mosseri$^{*\ 1}$
    \vspace{-0.2cm}
\end{tabular}
 \\ \\
\normalsize{$^1$Google DeepMind \quad \quad $^2$Tel Aviv University \quad \quad $^3$Weizmann Institute of Science \quad \quad $^4$Technion}
}

\begin{document}
\twocolumn[{%
\renewcommand\twocolumn[2][]{#1}%
\vspace{-1.3cm}	
\maketitle
\vspace{-0.7cm}	
\centering \centering
\includegraphics[width=0.9\textwidth]{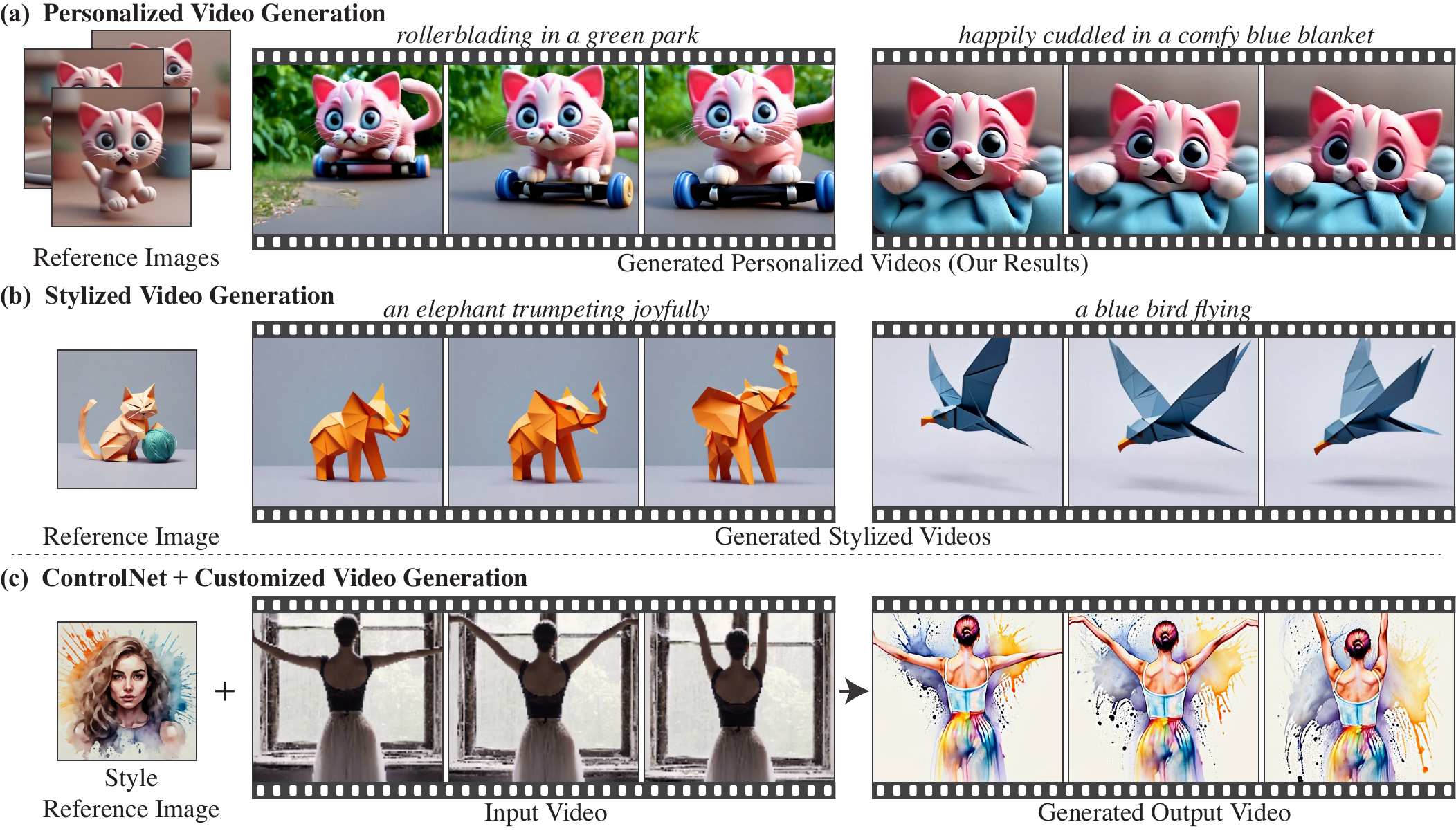}
\captionsetup[figure]{aboveskip=0.1cm}
\captionof{figure}
{\textit{Given a text-to-video (T2V) model built over a text-to-image (T2I) model, Still-Moving can adjust \emph{any} customized T2I weights to align with the T2V model. This is achieved by training lightweight adapters on \emph{still images}.
We demonstrate the effectiveness of our approach on a wide range of applications, including (a) personalized video generation, (b) stylized video generation, and (c) conditional video generation with ControlNet. In all cases, our method seamlessly combines the T2I model's spatial prior with the T2V model's temporal prior. Reference images are either from~\cite{avrahami2023chosen, hertz2023StyleAligned} or generated by Google.}}
\vspace*{0.12cm}
\label{fig:teaser}
}] 

\newcommand\blfootnote[1]{%
  \begingroup
  \renewcommand\thefootnote{}\footnote{#1}%
  \addtocounter{footnote}{-1}%
  \endgroup
}

\blfootnote{$^\dagger$ First author \quad $^*$ Core technical contribution \vspace{-0.2cm}} 
\blfootnote{\newline  Work was done while the first author was an intern at Google. \\
Webpage: \url{https://still-moving.github.io}}
\blfootnote{}

\vspace*{-0.3cm}
\begin{abstract}
\vspace*{-0.3cm}
Customizing text-to-image (T2I) models has seen tremendous progress recently, particularly in areas such as personalization, stylization, and conditional generation. However, expanding this progress to video generation is still in its infancy, primarily due to the lack of customized video data. 
In this work, we introduce Still-Moving, a novel generic framework for customizing a text-to-video (T2V) model, without requiring any customized video data. The framework applies to the prominent T2V design where the video model is built over a text-to-image (T2I) model (e.g., via inflation). We assume access to a customized version of the T2I model, trained only on still image data (e.g., using DreamBooth or StyleDrop).
Naively plugging in the weights of the customized T2I model into the T2V model often leads to significant artifacts or insufficient adherence to the customization data. 
To overcome this issue, we train lightweight \emph{Spatial Adapters} that adjust the features produced by the injected T2I layers.
Importantly, our adapters are trained on \emph{``frozen videos''} (i.e., repeated images), constructed from image samples generated by the customized T2I model. This training is facilitated by a novel \emph{Motion Adapter} module, which allows us to train on such static videos while preserving the motion prior of the video model. At test time, we remove the Motion Adapter modules and leave in only the trained Spatial Adapters. This restores the motion prior of the T2V model while adhering to the spatial prior of the customized T2I model.
We demonstrate the effectiveness of our approach on diverse tasks including personalized, stylized, and conditional generation. In all evaluated scenarios, our method seamlessly integrates the spatial prior of the customized T2I model with a motion prior supplied by the T2V model. 
\end{abstract}
    
\vspace{-0.1em}
\section{Introduction}
\label{sec:intro}

Given a small set of still images depicting a particular subject, humans can imagine how that subject would move and interact with its surroundings, even within completely different scenes and contexts. This capability is associated with the strong priors we have about dynamics, physics, and typical motions of objects in the world. Can a similar capability be obtained using a generative video model that has learned motion priors from broad video data?
In this paper, we address the task of customizing pre-trained text-to-video models to depict a specific subject or style, using only still image examples.

Our approach makes use of the tremendous progress that has been made in customizing text-to-image (T2I) models and aims to directly expand it to the realm of video generation. Specifically, we consider the prominent design where the text-to-video (T2V) model is built over a T2I model~\cite{lumiere,guo2023animatediff,gafni2022make}, and start by customizing the T2I model using existing approaches (e.g., DreamBooth~\cite{ruiz2022dreambooth} or StyleDrop~\cite{sohn2023styledrop}). This provides us with a T2I model that can generate novel image samples of the desired object or style. 

\begin{figure}
    \centering
    \setlength{\tabcolsep}{0.5pt}
    {\small
    \includegraphics[width=0.45\textwidth]{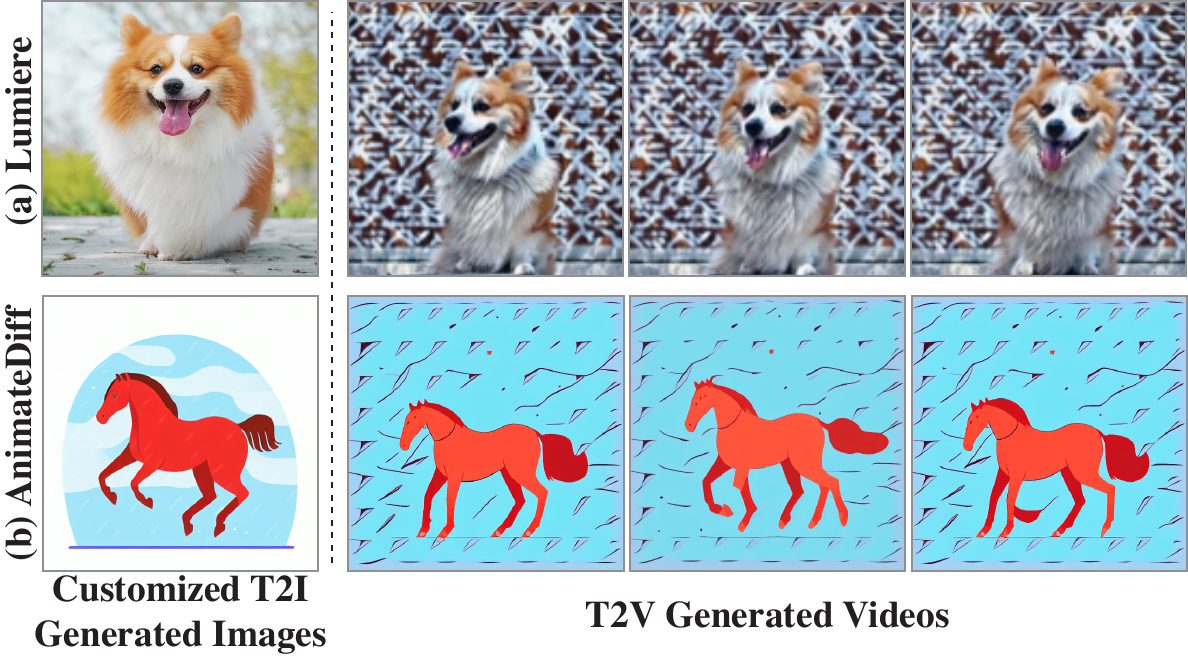}
    
    }
    \vspace{-6px}
    \caption{\textbf{Naive injection results}. \textit{Results of injecting customized T2I weights to a T2V model with (a) Lumiere~\cite{lumiere}, and (b) AnimateDiff~\cite{guo2023animatediff}. The injection often leads to significant artifacts (a), or fails to accurately preserve the features of the customized T2I model (b).}
    }
    \label{fig:motivation}
    \vspace{-12px}
\end{figure}

A naive approach to video customization is to simply replace the T2I weights within the T2V model with those of the customized T2I model~\cite{guo2023animatediff,liew2023magicedit}. However, the weights of the customized T2I model often deviate from their counterparts within the inflated T2V model, which leads to a mismatch in feature distributions. Therefore, as recently observed in several works~\cite{lumiere,singer2024video}, this approach can result in significant artifacts or low fidelity to the customization data (see Fig.~\ref{fig:motivation}). 
In this work, we propose a generic framework for harnessing the generative 2D prior of the customized image model, while preserving the motion prior of the pre-trained T2V model. 

Ideally, we would want to fine-tune the T2V model by providing direct supervision on its output frames. However, we do not have access to any video data of the customized content. Our key idea to overcoming this challenge is to fine-tune the T2V model on ``frozen videos'' that are constructed by temporally duplicating still images generated by the customized T2I model.  A pivotal challenge in this approach is to preserve the motion prior of the T2V model and avoid teaching it to generate static content. We tackle this challenge in two stages. We first train Motion Adapters, which are lightweight residuals to the temporal attention layers that cause the T2V model to generate static videos. We then use these Motion Adapters as motion switches: we turn them on to allow us to fine-tune the T2V model on frozen customized videos, and then remove them to restore the model's motion prior. The T2V fine-tuning is achieved by injecting the customized T2I weights and training only lightweight Spatial Adapters to amend the mismatch in feature distribution. 

We showcase our approach on two prominent inflated T2V models -- Lumiere \cite{lumiere} and AnimateDiff \cite{guo2023animatediff}. We explore a diverse set of customization tasks, including personalized generation, stylized generation, and conditional generation. As illustrated in Fig.~\ref{fig:teaser}, in all cases our method manages to non-trivially combine the spatial prior from the customized T2I model with a unique motion prior provided by the T2V model. For example, the plasticine cat in Fig.~\ref{fig:teaser}(a) demonstrates our method's ability to couple objects with both realistic motion (left) and intricate facial expressions (right). Figures~\ref{fig:teaser}(b),(c) demonstrate an ability to creatively match a reference style with unique motion. Notably, even when coupled with the restrictive ControlNet  condition~\cite{zhang2023adding}, our method  produces additional splashing motion in the background behind the subject, which fits the input reference style image, and does not appear in the conditioning video (Fig.~\ref{fig:teaser}(c)).

Through extensive experiments, we demonstrate our method's ability to inject the spatial prior of a customized T2I model into a T2V model while benefiting from the respective rich priors of both models. We show that our framework is superior to existing approaches, which do not pay special attention to the temporal layers. Our framework is generic, lightweight, and can be applied to any T2V model that is built on top of a T2I model.

\section{Related Work}
\begin{figure*}
\centering
\includegraphics[width=\textwidth]{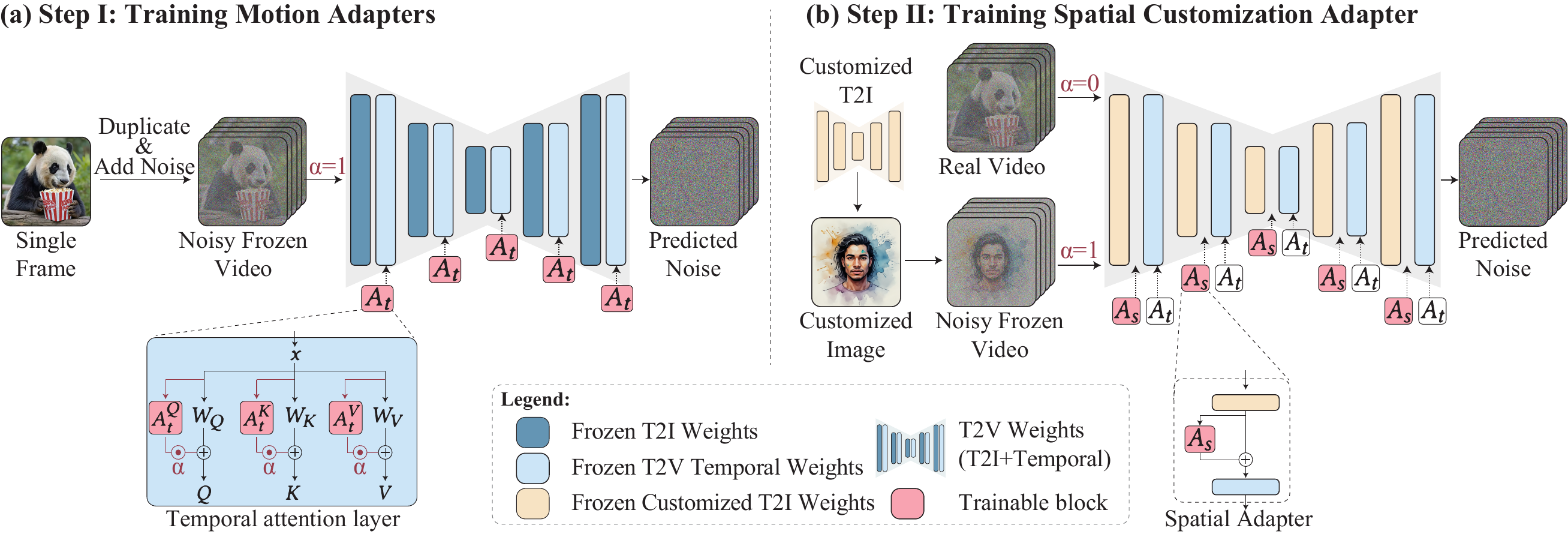}
    \vspace{-9px}
    \caption{\textbf{Still-Moving pipeline.} 
    \textit{Given a T2V model inflated from a T2I model, and a customized version of the T2I model (e.g., fine-tuned using DreamBooth, StyleDrop, etc.), we inject and adapt the customized T2I weights to the T2V model in two steps. (a) We introduce Motion Adapters that control the level of motion/dynamics in videos generated by the model. The Motion Adapters are implemented as LoRA layers on top of the temporal attention blocks and enable training on ``frozen'' customized videos (videos composed of a duplicated single customized frame, using a scale of $\alpha=1$, see Section~\ref{sec:method}). Then, (b) we inject the customized T2I weights and train Spatial Adapters on a combination of customized images (with $\alpha=1$) and natural videos (with $\alpha=0$).}}
    \label{fig:architecture}
    \vspace{-14px}
\end{figure*}

\paragraph{\textbf{Text-to-Image Customization}}

We focus on three prominent customization tasks, namely \emph{personalization}, \emph{stylized generation}, and \emph{conditional generation}.
\emph{Personalization} aims to introduce a new concept to the model and allow it to generate novel scenes containing the concept. A dominant approach to solving this task is optimizing some or all of the model's weights to reconstruct a small set of concept images~\cite{ruiz2022dreambooth,ruiz2023hyperdreambooth,kumari2022customdiffusion, han2023svdiff, hu2022lora,jiang2023videobooth}. 
Other approaches include token optimization~\cite{gal2022textual,voynov2023p+,alaluf2023neural,conceptor,palp}, encoder-based personalization~\cite{gal2023encoder,arar2023domain}, and training-free personalization~\cite{ye2023ip-adapter,wang2024instantid,wei2023elite,shi2023instantbooth}.

\emph{Stylized generation} is tasked with distilling appearance-based features (e.g., colors, textures, lighting) of a reference input to the generations by the model. Similar to personalization, weight optimization is a common approach to obtaining stylized generation~\cite{sohn2023styledrop,ruiz2022dreambooth,dblora,hu2022lora,dora}.

Finally, \emph{conditional generation} introduces additional explicit control to the generation process. The most common approach to do so, dubbed ControlNet~\cite{zhang2023adding}, is based on replicating the diffusion block weights in a residual branch while training only the weights of the added branch. 
This approach has recently shown promise for video generation models as well~\cite{liew2023magicedit}.

\vspace*{-1em}
\paragraph{\textbf{Text-to-Video Customization}}
In this work, we focus on the prominent ``inflation'' approach in which a T2I model is  ``inflated'' into a video model by adding temporal blocks between the existing spatial blocks \cite{singer2022make,blattmann2023videoldm,girdhar2023emu,ge2023preserve,guo2023animatediff,Yuan_2024_WACV}. 
The dominant approach for video customization is relying on inflated models, and replacing their T2I weights with customized weights~\cite{liew2023magicedit,guo2023animatediff}, or masking out the temporal layers and applying existing T2I methods on the spatial layers~\cite{molad2023dreamix}. 
As demonstrated in recent works~\cite{lumiere,singer2024video}, this approach leads to low fidelity to the customized data or noticeable artifacts. 
In alignment with these works, we find that even AnimateDiff~\cite{guo2023animatediff}, which demonstrated promising weight replacement results, falls short of adhering to the customized content when applying detailed styles or personalized subjects (see Fig.~\ref{fig:motivation}(b)). 

Related to our task, VideoBooth~\cite{jiang2023videobooth} conditions video generation on content extracted from a single image. Specifically, the object in the input image is segmented and injected into the model via cross-attention and cross-frame attention modules. This approach is restricted to the information provided in the input image and struggles to generalize across scenes and to object dynamics that differ from the input (see SM). Furthermore, this method does not support additional tasks (e.g., stylized generation) as we do. 
Other works~\cite{guo2023animatediff,2023videocomposer,he2024cameractrl,yang2024direct} propose motion customization methods, and combine them with personalized subjects~\cite{materzynska2023customizing,Wei_2024_CVPR}.

Conceptually closest to ours (in the sense of combining customized T2I weights with inflated temporal weights) is the concurrent work EVE~\cite{singer2024video}, which targets text-based video editing by distilling knowledge from both an image editor and a T2V model. However, their approach requires a complex system of discriminators for each customized T2I model, and for the base T2V model. Additionally, it involves a distillation loss through multiple diffusion steps from both models, making it very computationally expensive. In contrast, our framework aims to enable the simple plug-and-play approach similar to~\cite{guo2023animatediff} with \emph{minimal, lightweight adaptation} that is \emph{generic} and applicable to different T2V inflation approaches and applications.

\section{Method}
\label{sec:method}
Given a video model $V$ inflated from a T2I model $M$, and a customized T2I model $M'$ fine-tuned from $M$, our goal is to enable a ``plug-and-play'' injection of the customized weights $M'$ into $V$. 
As discussed in~\cref{sec:intro}, the naive approach of simply substituting $M$ with $M'$ leads to unsatisfactory results due to the shift in feature distribution caused by the switch (\cref{fig:motivation}).
To mitigate this shift, we propose training lightweight (in both the number of parameters and the number of optimization steps) \emph{Spatial Adapters}, which are tasked with projecting the activations at the outputs of the injected layers from $M'$ back to the distribution of the temporal layers of $V$. 
A major challenge with this approach is that the adapters should be trained on \emph{customized video data}, which is often unattainable. Note that while still images can be duplicated to form (frozen) videos, naively training on such frozen videos leads to a loss of motion generation abilities (see~\cref{sec:ablations}). To overcome this, we propose to employ novel \emph{Motion Adapters} to enable training on \emph{image data} without losing the model's motion prior.
In the following, we briefly describe the inflation approach, and then detail the two main components of our method, namely the Motion Adapters and the Spatial Adapters.

\subsection{Preliminaries}

We focus on the prominent ``inflation'' paradigm, where a T2V diffusion model $V$ is an ``inflated'' variant of a T2I model $M$, obtained by weaving temporal blocks between the spatial T2I blocks. 
An important component of inflated models is temporal attention~\cite{singer2022make,guo2023animatediff}, which is used to share information across frames. Specifically, denote the input sequence by $X\in \mathbb{R}^{F\times H \times W \times C}$, where $F$ is the number of video frames, $H, W$ are the frame spatial dimensions, and $C$ is the channel dimension. Temporal attention operates as a vanilla self-attention block over the reshaped sequence $X\in \mathbb{R}^{H \cdot W\times F \times C}$. First, the input sequence is projected into queries, keys, and values, using projection matrices $W_Q,W_K,W_V$, leading to
\begin{align}
    Q \in \mathbb{R}^{H \cdot W\times F \times d_k},
    K \in \mathbb{R}^{H \cdot W\times F \times d_k},
    V \in \mathbb{R}^{H \cdot W\times F \times d_k}, 
    \label{eq:projection}
\end{align}
where $d_k$ is the attention embedding dimension. Next, an attention matrix is calculated as
\begin{align}
    A = \text{softmax}\left(\frac{Q \cdot K^T}{\sqrt{d_k}}\right),
\end{align}
where $A \in \mathbb{R}^{H\cdot W \times F \times F}$. 
Finally, the output of the block is calculated as $Y = A \cdot V$. 
In words, the temporal attention mechanism splits the video into $H \cdot W$ temporal ``needles'' of dimension $F$, and performs the attention operation for each such ``needle''. 
This allows the model to share information across frames, which is crucial in determining the motion in the resulting video. Therefore, the temporal attention blocks are a natural candidate for placing our Motion Adapters. 

\begin{figure*}
    \centering
    \setlength{\tabcolsep}{0.5pt}
    \includegraphics[width=\textwidth]{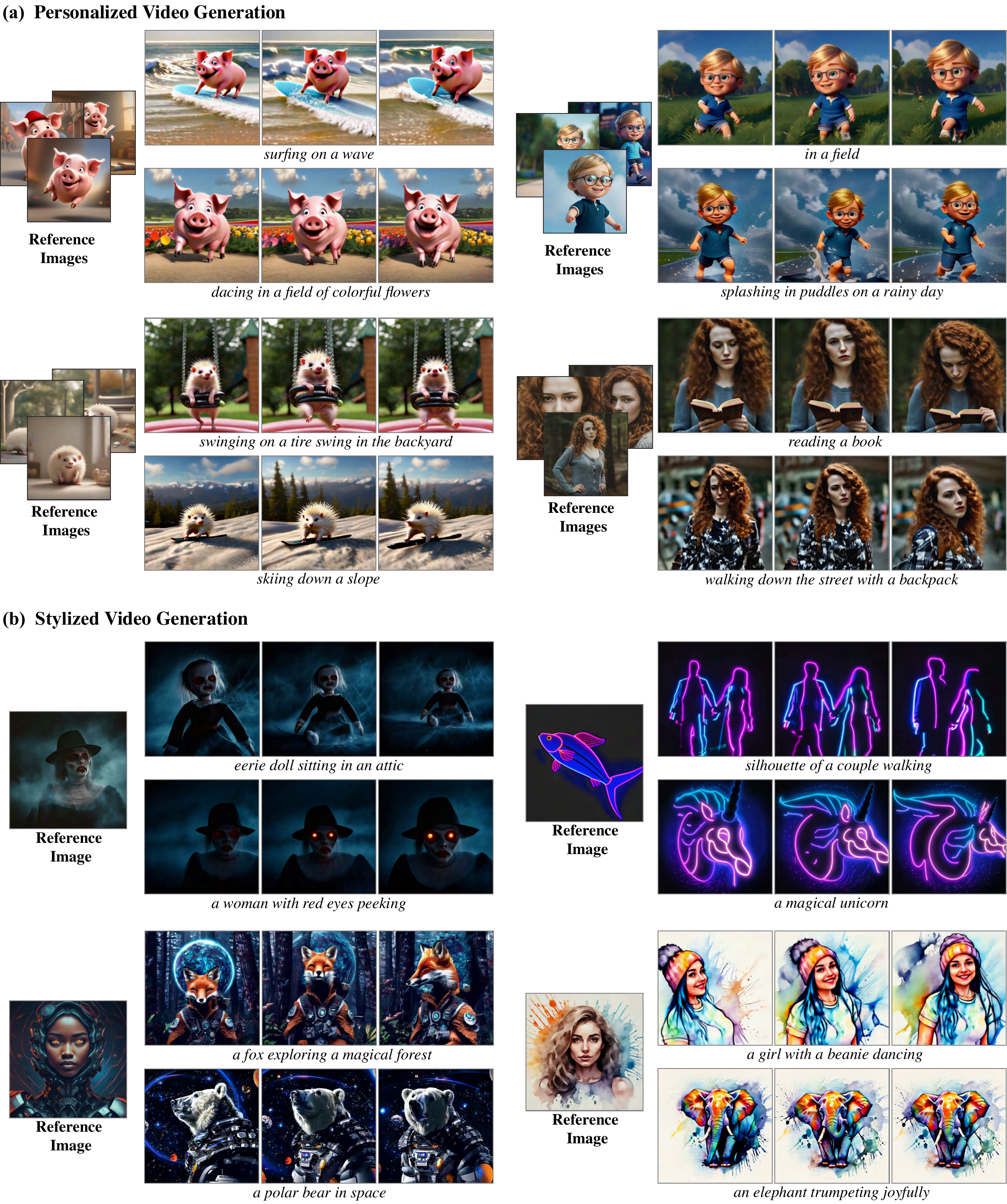}
    \vspace{-8px}
    \caption{\textbf{Qualitative results.} \textit{Examples of applying Still-Moving for (a) personalized video generation and (b) stylized video generation over Lumiere~\cite{lumiere}. Our method maintains the spatial prior of the customized T2I model while incorporating corresponding motion derived from the T2V prior. Reference images are either from~\cite{avrahami2023chosen} or generated by Google.
    }}
    \vspace{-6px}
    \label{fig:qualitative}
\end{figure*}

\subsection{Motion Adapters}

A key component of our method is the ability to train the weights of $V$ on frozen image data. To accomplish this goal without introducing out-of-distribution input, we propose training lightweight Motion Adapters to control the existence of motion in the videos generated by the model. The Motion Adapters are trained once over the vanilla, non-customized T2V model. 
Our implementation is based on a Low-Rank Adaptation (LoRA)~\cite{hu2022lora,dblora} of the temporal attention projection matrices (see~\cref{fig:architecture}(a)) as
\begin{align}
    \tilde{W} =  W + \alpha \cdot A^{W,\text{down}}_{t} \cdot A^{W,\text{up}}_{t} \,
\end{align}
for all $W \in \{W_Q, W_K, W_V\}$. Here $\alpha$ is the adapter scale and $A^{W,\text{down}}_{t} \in \mathbb{R}^{C \times r}$, $\smash{A^{W,\text{up}}_{t} \in \mathbb{R}^{r \times d_k}}$ are the adapter matrices. 
The matrices $\smash{A^{W,\text{down}}_{t}}$ are initialized with random values, while $\smash{A^{W,\text{up}}_{t}}$ are initialized to zeros, such that before training, the model is equivalent to $V$.

During the training, we set $\alpha=1$ and employ a small set of videos from the model's training set (see SM for more details). For each video, we randomly select a single frame, duplicate it $F$ times, and train the adapter layers with the diffusion denoising objective. In other words, the Motion Adapters are trained to generate frozen videos depicting a duplicated random frame from the video model's distribution (see~\cref{fig:architecture}(a)). Once trained, the Motion Adapters enable training the video model weights on frozen data, as we detail in Sec.~\ref{sec:spatial_adapters}. Note that by setting $\alpha=0$, the model retains its ability to generate motion. 

Interestingly, we find that the Motion Adapters' ability to control the amount of motion generalizes to negative scales. For example, when setting $\alpha = -1$, the model produces videos with an \emph{increased} amount of motion (see SM). Therefore, besides allowing to customize the T2V model on images, the Motion Adapters may be used during inference to control the amount of motion in the generated videos.

\begin{figure*}
    \centering
    \setlength{\tabcolsep}{0.5pt}
    {\small
    \includegraphics[width=0.95\linewidth]{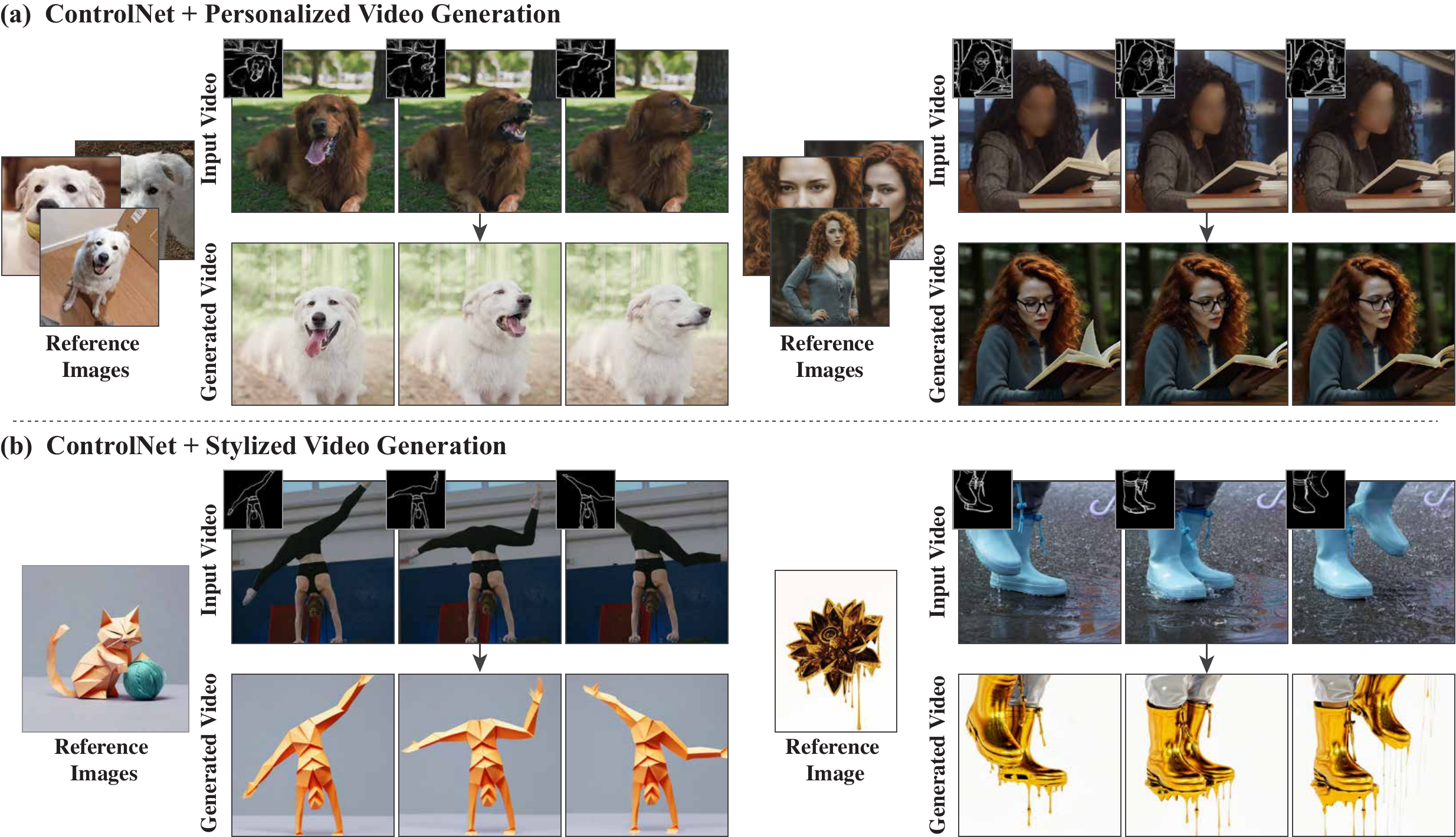}
    
    }
    \vspace{-6px}
    \caption{\textbf{Conditional generation with Still-Moving.} \textit{We present results of combining our method with ControlNet~\cite{zhang2023adding} for (a) \text{conditional personalized generation} and (b) \text{conditional stylized generation}. The reference images of the woman and the cat are generated images, provided by Google and from \cite{hertz2023StyleAligned}, respectively,  whereas the reference images of the dog and the flower are real.}}
    \vspace{-10px}
    \label{fig:controlnet}
\end{figure*}

\subsection{Spatial Adapters}
\label{sec:spatial_adapters}
At the heart of our method lies the premise that the deviation in feature distribution can be fixed using a simple linear projection. We, therefore, propose to add Spatial Adapters after every injected customized T2I layer (see~\cref{fig:architecture}(b)). These adapters are tasked with amending the distribution gap at the input to the temporal layers. Therefore, their training necessitates propagating gradients through the entire video model. Similarly to the Motion Adapters, the Spatial Adapters are implemented as a multiplication of low-rank matrices
\begin{align}
    \text{Adapter} (X) = X \cdot  \left( I + A^{\text{down}}_{s} \cdot A^{\text{up}}_{s}\right),
\end{align}
where $A^{\text{down}}_{s} \in \mathbb{R}^{C \times r}, A^{\text{up}}_{s} \in \mathbb{R}^{r \times C}$ are the layer-specific low-rank adapter matrices. Similarly to the Motion Adapters, $A^{\text{down}}_{s}, A^{\text{up}}_{s}$ are initialized randomly and with zeros respectively, such that before training the model is equivalent to $V$. We train the Spatial Adapters with the diffusion reconstruction loss using a combination of images and videos. Specifically, we use images generated by the customized T2I model $M'$, and $40$ non-customized videos to preserve the prior of the inflated model, following DreamBooth~\cite{ruiz2022dreambooth} (see~\cref{fig:architecture}(b)). We employ the Motion Adapters with a scale of $\alpha=1$ for the frozen videos, and $\alpha=0$ for the prior preservation videos.

\section{Experiments}
We present extensive qualitative and quantitative evaluation, and  compare our method with prominent baselines. We perform evaluation on the potent customization tasks of personalization using DreamBooth~\cite{ruiz2022dreambooth} and stylization using StyleDrop~\cite{sohn2023styledrop}. Additionally, we demonstrate the combination of our method with ControlNet~\cite{zhang2023adding}, which facilitates the customization of existing videos to a personalized subject or a given style while preserving their original structure. 
We present results on two notable inflated T2V models, namely, Lumeire~\cite{lumiere} built on Imagen~\cite{ho2022imagen}, and AnimateDiff~\cite{guo2023animatediff} built on Stable Diffusion~\cite{rombach2021highresolution}, to demonstrate the robustness of our method. For more implementation details of our method for both models see the SM document.

\begin{figure*}
    \centering
    \setlength{\tabcolsep}{0.5pt}
    {\small
    \includegraphics[width=0.97\textwidth]{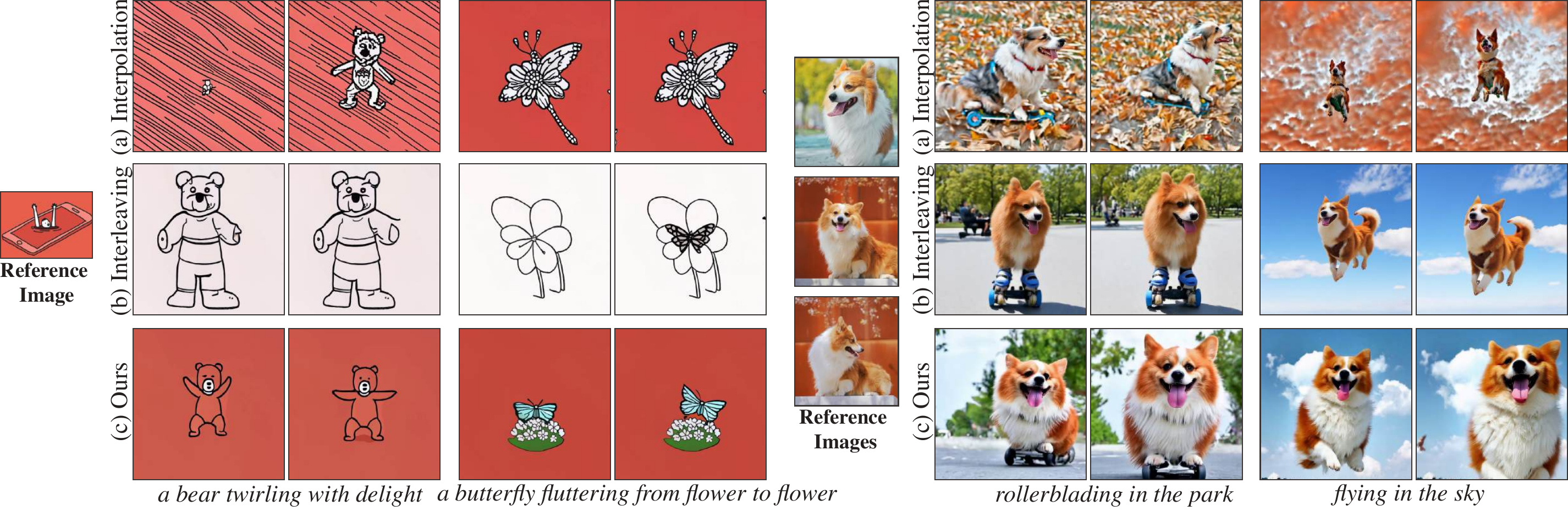}
    
    }
    \vspace{-6px}
    \caption{\textbf{Qualitative comparison.} \textit{We present qualitative comparisons to the leading baselines, interpolation, and interleaved training. Interpolation often fails to mitigate all the artifacts, and results in degraded character fidelity. Interleaved training falls short of capturing all the features in the customized data. Reference images are real images from~\cite{ruiz2022dreambooth,sohn2023styledrop}.}
    }
    \vspace{-14px}
    \label{fig:comparison}
\end{figure*}

\paragraph{Baselines} We consider three prominent baselines: (1) \emph{Naive weight injection:} replacing the T2I model weights in the T2V model by the customized T2I weights, as applied by~\citet{guo2023animatediff,liew2023magicedit},  (2) \emph{weight interpolation:} interpolating between the customized weights and the pre-trained T2I weights, as applied by~\citet{lumiere}, and (3) \emph{interleaved training:} inspired by a common approach to T2V training~\cite{gafni2022make,sora}, we fine-tune the T2I weights in the video model using: (i) customized images where the temporal layers are disabled, and (ii) natural videos where the temporal layers are enabled. The crucial difference between this baseline and our method is that the final generated videos are not explicitly supervised using the customized image data, as in our framework. 

\paragraph{Evaluation Data} We construct a diverse dataset containing both personalized objects and styles. For personalization, we include five different objects from the DreamBooth dataset~\cite{ruiz2022dreambooth}, including realistic (e.g., dogs, cats) and out-of-distribution (e.g., toys) ones. For the stylization evaluation, we use the three styles presented in Lumiere~\cite{lumiere} in addition to two challenging realistic and highly detailed styles presented in~\cref{fig:qualitative}. Each such style enables a comprehensive evaluation of a wide variety of prompts encompassing different scenes and objects.
This results in a dataset of $10$ different personalized objects and styles. 
For each item in our data, we generate $10$ comparison videos on diverse prompts with the same random seed, amounting to $100$ comparison videos per baseline. Please refer to the SM for more details on our dataset.

\begin{table}
\resizebox{\columnwidth}{!}{
\centering
\begin{tabular}{lcccc}
\toprule[1.5pt]
& \multicolumn{2}{c}{\bf Personalization} & \multicolumn{2}{c}{\bf Stylization} \\
Method & CLIP-I & CLIP-T & CLIP-I & CLIP-T \\
\midrule
    Injection     & 0.680 \scriptsize $\pm$ 0.018    & 0.236 \scriptsize $\pm$ 0.007     & 0.670  \scriptsize $\pm$ 0.013   & 0.211 \scriptsize $\pm$ 0.007 \\
    Interpolation & 0.724 \scriptsize $\pm$ 0.009    & 0.320 \scriptsize $\pm$ 0.006     & 0.664 \scriptsize $\pm$ 0.008    & 0.259 \scriptsize $\pm$ 0.005\\
    Interleaving  & 0.717 \scriptsize $\pm$ 0.011    & 0.324 \scriptsize $\pm$ 0.006     & 0.642 \scriptsize $\pm$ 0.011    & 0.275 \scriptsize $\pm$ 0.008 \\
    Ours          & {\bf0.772} \scriptsize $\pm$ 0.007 & {\bf0.325}  \scriptsize $\pm$ 0.004 & {\bf0.673} \scriptsize $\pm$ 0.01 & {\bf0.297} \scriptsize $\pm$ 0.005 \\
\bottomrule[1.5pt] \\    
\end{tabular}
}
\vspace{-13px}
\caption{{\bf Quantitative metrics.}
\textit{Evaluation metrics using CLIP for image-image and text-image similarity to measure subject / style fidelity (CLIP-I) and prompt alignment (CLIP-T), respectively.}
}
\label{tab:metrics}
\vspace{-18px}
\end{table}

\subsection{Qualitative Results}
\begin{figure*}
    \centering
    {\small
    \includegraphics[width=0.9\textwidth]{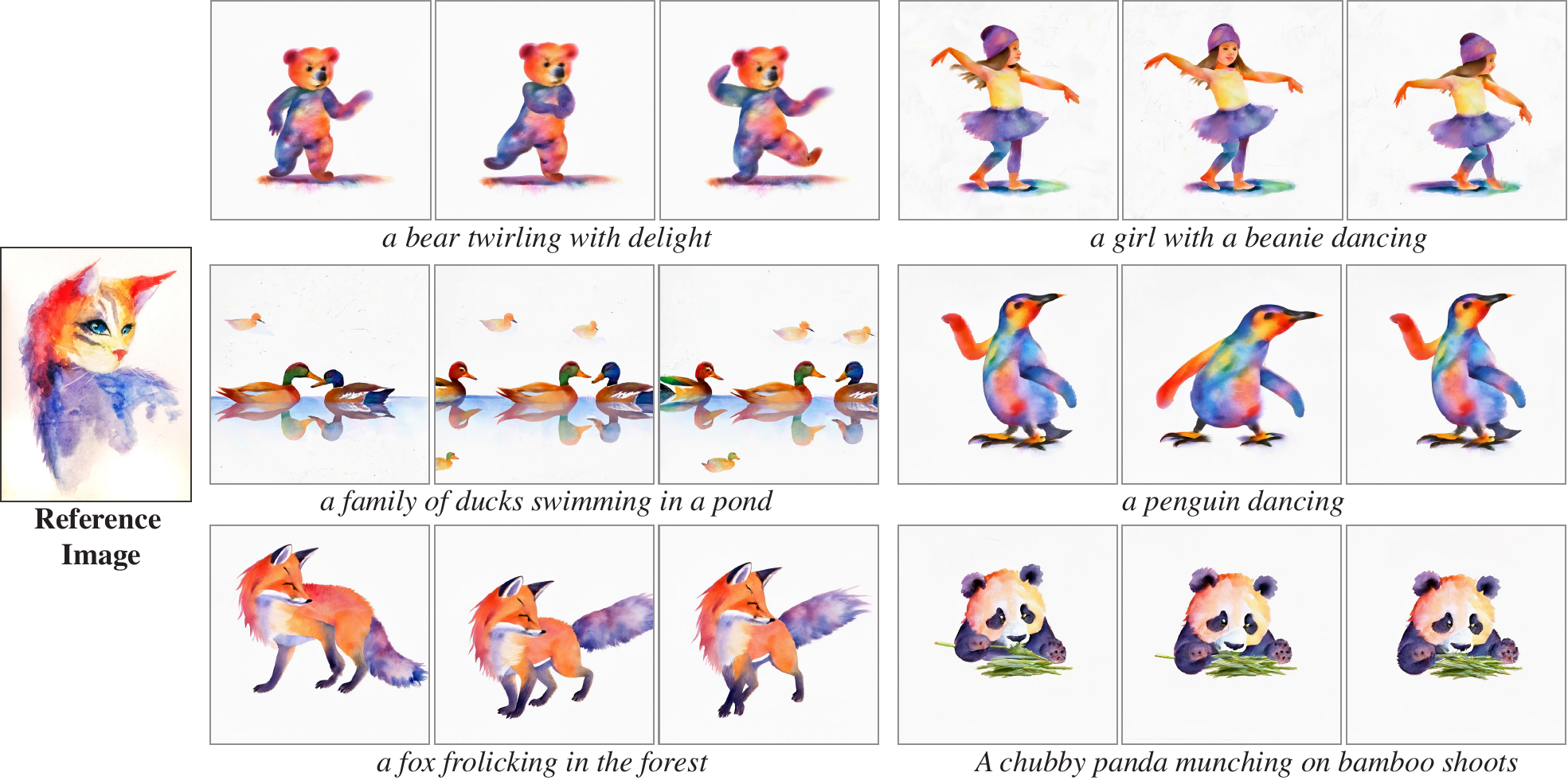}
    
    }
    \vspace{-8px}
    \caption{\textbf{Uncurated results.} \textit{We present uncurated results (obtained with a single random seed without any manual selection) of Still-Moving on Lumiere~\cite{lumiere}. The prompts used were extracted directly from Lumiere's stylized generation samples. Reference image is not generated; see attribution page for details.}}
    \vspace{-12px}
    \label{fig:uncurated}
\end{figure*}

Figures~\ref{fig:teaser},\ref{fig:qualitative} and \ref{fig:uncurated} present results obtained by our method with Lumiere for both video personalization and stylization, on a variety of challenging customized T2I models. The personalization characters are either from~\citet{avrahami2023chosen} or generated by Google. Similarly, the style references are either from~\citet{sohn2023styledrop, hertz2023StyleAligned} or generated by Google. As can be observed, our method remains faithful to the spatial prior of $M'$, while accompanying it by matching creative motion priors from $V$. For example, 
the ``watercolor splash'' style (Fig.~\ref{fig:qualitative}(b)) is accompanied by splashing color motion, 
the ``horror movie'' style (Fig.~\ref{fig:qualitative}(b)) incorporates fog motion, etc. For personalization, our method demonstrates an innate ability to animate both realistic (e.g., the woman in Fig.~\ref{fig:qualitative}(a)) and animated characters (e.g., the cat in Fig.~\ref{fig:teaser}), while producing diverse backgrounds, scenes, and dynamics (e.g., surfing, dancing).

\begin{figure}
    \centering
    {\small
    \includegraphics[width=0.48\textwidth]{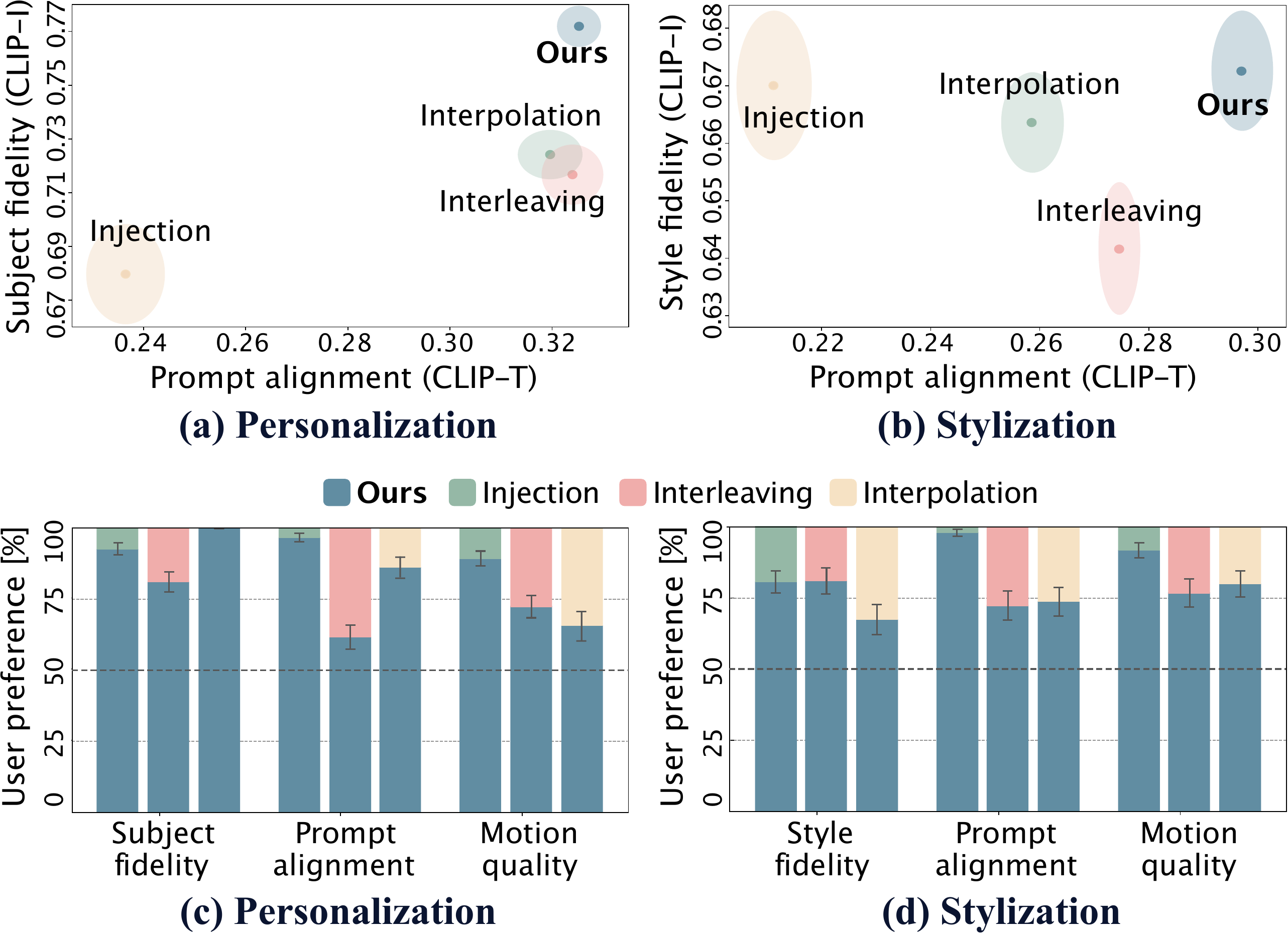}
    
    }
    \vspace{-4px}
    \caption{\textbf{Quantitative evaluation plots and user study.} \textit{We compare our method to each of the baselines using both automatic metrics (CLIP similarity) and a user study. Subplots (a) and (b) correspond to the results of~\cref{tab:metrics} for personalization and stylization, repectively. Subplots (c) and (d) visualize the percentage of user votes in our favor vs those that favor the baselines. Our method was preferred by users across all measured categories.}}
    \label{fig:metrics_plots}
    \vspace{-16px}
\end{figure}

Figure~\ref{fig:animatediff} presents a qualitative comparison between our method and the naive injection baseline on AnimateDiff~\cite{guo2023animatediff}. Importantly, the authors of AnimateDiff highlight the model's ability to seamlessly plug in fine-tuned T2I weights. As can be observed, this approach often produces unsatisfactory results for intricate customized data. For stylization, the ``melting golden'' style (top row) displays a distorted background and lacks the melting drops that are characteristic to the style. 
For personalization, the chipmunk's features are not captured accurately (e.g., the chicks and the forehead's color). Additionally, the identity of the chipmunk changes across the frames. In contrast, when applying our method, the ``melting golden'' background matches the reference image and the model produces dripping motion. Similarly, the chipmunk maintains a consistent identity that matches the reference images.

Next, Figs.~\ref{fig:teaser} and~\ref{fig:controlnet} present results of combining our method with ControlNet. Given a driving video, our method enables the generation of a customized video that follows its structure while incorporating a style or character. For example, the video in the top row of Fig.~\ref{fig:controlnet} is transformed to display the reference woman, while maintaining the main characteristics of the driving video.  
Interestingly, the combination of ControlNet with Still-Moving can also incorporate dynamic motion derived from the style, in addition to the motion dictated by the condition. For example, the ``melting golden'' style in Fig.~\ref{fig:controlnet} adds dripping motion to the scene.

Figure~\ref{fig:comparison} presents a qualitative comparison with the leading baselines, namely \emph{interpolation} and \emph{interleaved training}, on examples from our evaluation dataset. Observe that the interpolation baseline inherently loses information from $M'$ in the process of the interpolation. This is most evident in the personalization example in~\cref{fig:comparison}, where the identity of the dog is lost in the interpolation. Additionally, observe that the interpolation is not always sufficient to overcome the distribution gap, and significant artifacts may remain. 
The interleaved training baseline falls short of capturing the reference images accurately in both cases. This can be attributed to the fact that the baseline employs the customized data to modify the T2I layers of the model while ignoring the temporal layers. As shown in~\cref{fig:motivation}, the combination of the fine-tuned spatial layers and the pre-trained temporal layers often causes a distribution shift, which leads to a lack of adherence to the customized data. 
Kindly refer to the supplemental website for many more video results.

\begin{figure}
    \centering
    \vspace{-2px}
    {\small
    \includegraphics[width=0.43\textwidth]{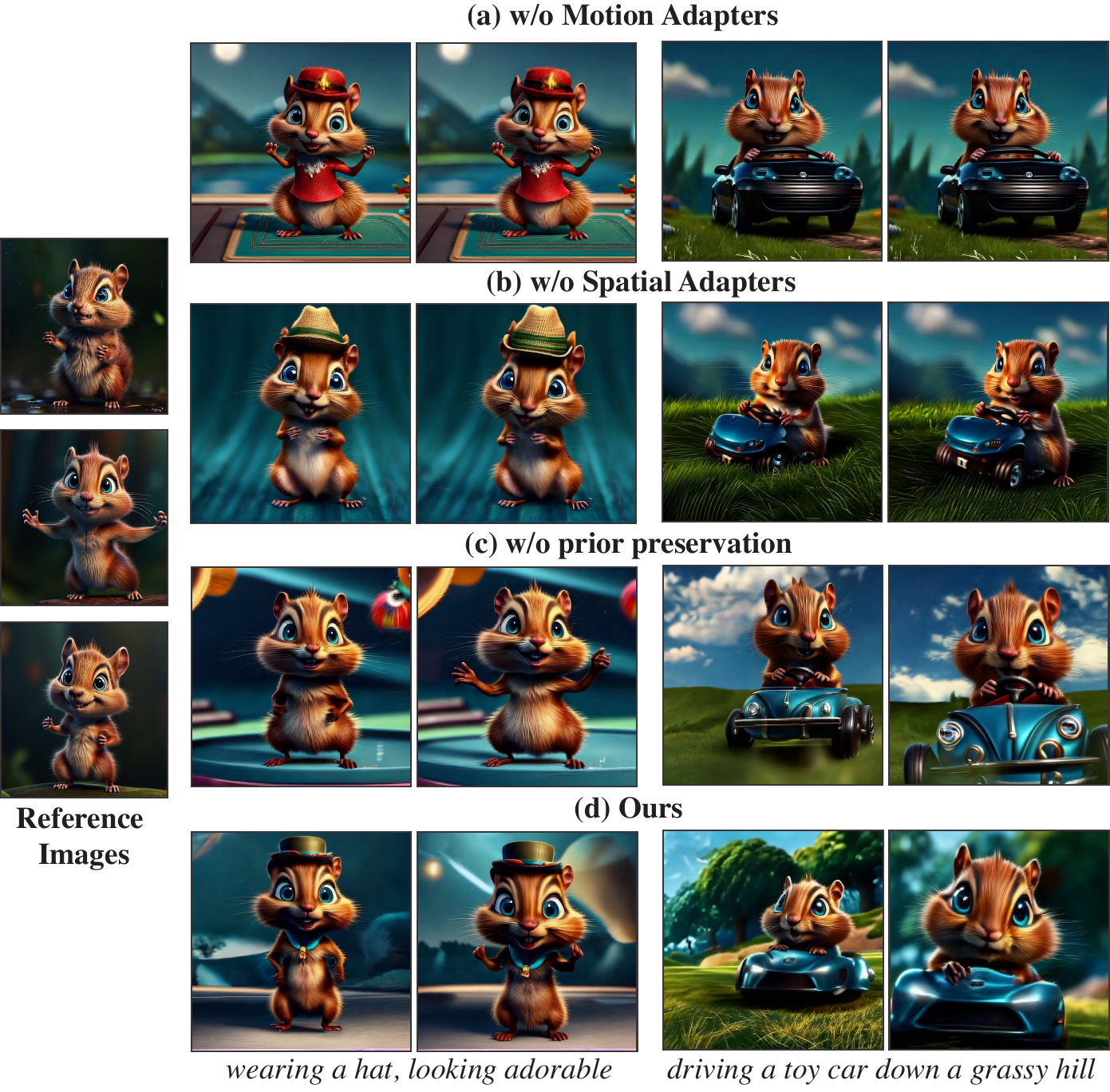}
    }
    \vspace{-6px}
    \caption{\textbf{Ablation study.} \textit{We ablate the three key components of our method, namely (a) the  Motion Adapters, (b) the Spatial Adapters, and (c) the prior preservation loss.  When removing the Motion Adapters, the model loses its motion prior. Without the Spatial Adapters we observe an overfit, and the prior preservation removal harms the diversity and the amount of motion. Reference images were generated by Google.}
    }
    \label{fig:ablations}
    \vspace{-15px}
\end{figure}

\subsection{Quantitative Results} Next, we present a quantitative comparison with the baselines through automatic metrics and a user study.

\emph{Metrics.} Following common practice in similar works from the image customization domain~\cite{ruiz2022dreambooth,sohn2023styledrop}, we consider the CLIP~\cite{radford2021learning} text-image similarity w.r.t. the driving prompt per frame (CLIP-T), and the image-image similarity w.r.t. the customized data per frame (CLIP-I).
The results of the automatic evaluation on both subsets of our dataset (personalization, stylization) are reported in~\cref{tab:metrics} and~\cref{fig:metrics_plots}(a), (b). Our method outperforms all baselines across all the examined subsets and all examined criteria by a statistically significant margin, as observed by the standard error of mean (SME) reported in~\cref{fig:metrics_plots} and~\cref{tab:metrics}. 
The trends in~\cref{fig:metrics_plots} are similar to those observed in our qualitative comparisons (\cref{fig:comparison}). The interpolation and injection baselines perform better on stylization, which can be attributed to the larger distribution gap for personalized T2I models. Conversely, the interleaved baseline is unable to adapt to novel styles due to the fact that it trains only the T2I model weights on customized data.

It is important to note that it is possible to reach very high CLIP-I and CLIP-T scores by simply generating a single image with the customized T2I model and duplicating it to form a frozen video. 
In the absence of a motion metric to evaluate the quality of the generated motion (as pointed out by multiple works ~\cite{lumiere,girdhar2023emu,ho2022imagen}), we resort to a user study to further evaluate the quality of the generated videos.
Our study was conducted with 31 participants, each was tasked with voting on 20 comparisons containing 3 questions. This resulted in a total of 1860 votes. In each comparison, the participants were shown two videos randomly sampled from the comparison dataset- one by our method and the other by a baseline, both depicting the same reference subject/style and generated with the same prompt and seed. Participants were asked to answer 3 questions: (1) ``Which video better follows the provided reference image?'', (2) ``Which video is more aligned with the provided text prompt?'', and (3) ``Which video displays better motion?''. As shown in~\cref{fig:metrics_plots}(c),(d), the user study results are consistent with the automatic metrics in terms of subject fidelity and prompt alignment. Additionally, the study demonstrates that our method produces videos with superior motion quality compared to the baselines by a statistically significant margin, as indicated by the SME.

\vspace{-3px}
\subsection{Ablation Study}
\label{sec:ablations}
\vspace{-3px}

We ablate the three main design choices of our method, namely (i)~applying the Motion Adapters, (ii)~applying the Spatial Adapters, and (iii)~using a prior preservation loss. The results of these ablations are presented in~\cref{fig:ablations}. When removing the Motion Adapters and training on static videos, the model produces nearly static videos (top row). When removing the Spatial Adapters, and instead training \emph{all} the network weights, we observe that the motion is still significantly reduced (second row), and the background is less diverse. This can be attributed to the significant increase in the number of optimized parameters, which allows the model to overfit easily, even in the presence of the Motion Adapters. These two ablations demonstrate that both of the main novel components of our method are necessary to obtain results with meaningful motion. Finally, when removing the prior preservation component (third row in~\cref{fig:ablations}), the model loses some of its ability to generalize, for example, in the first column of~\cref{fig:ablations}, the chipmunk is not wearing a hat. Additionally, we find that the prior preservation helps the model maintain its motion prior better, and without it, the motion is somewhat reduced.

\subsection{Limitations}

\begin{figure}
    \centering
    {\small
    \includegraphics[width=0.46\textwidth]{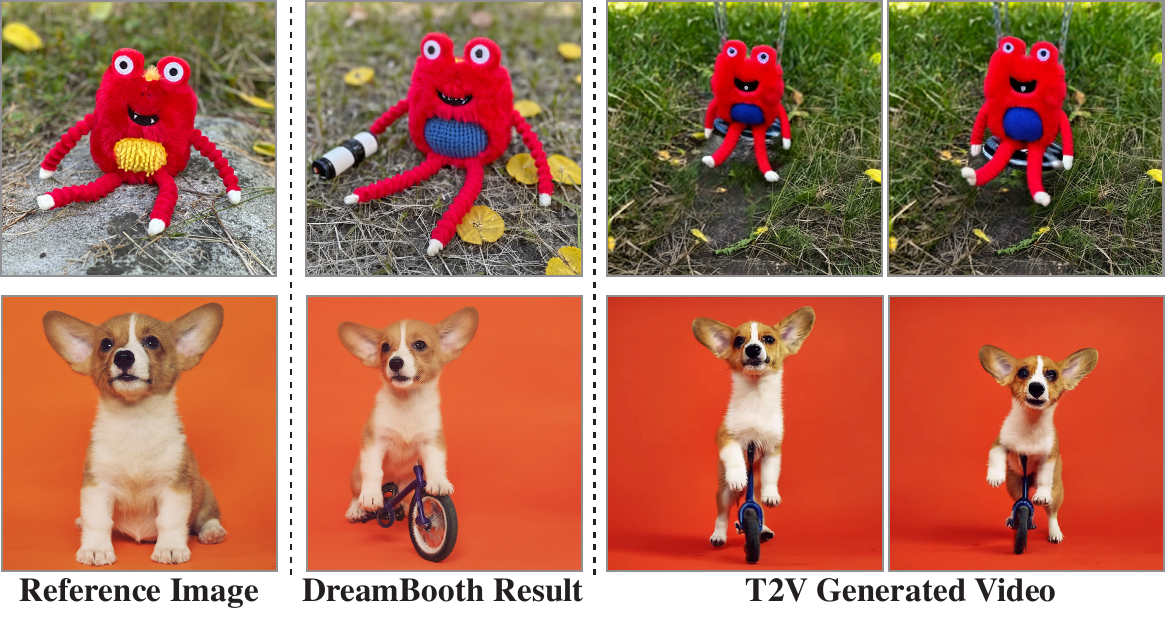}
    
    }
    \vspace{-6px}
    \caption{\textbf{Limitations.} \textit{Our method is  limited by the quality of the injected T2I model, this can cause inaccurate identity (top) and overfitting (bottom). The reference images are not generated; see attribution page for details.}
    }
    \label{fig:limitations}
    \vspace{-18px}
\end{figure}

Our work  enables a plug-and-play injection of customized T2I weights to a T2V model. Thus, our results are inherently limited by those of the customized T2I model. For example,  when the T2I model fails to accurately capture some feature of the customized object, the videos produced by our model may display a similar behavior. Similarly, if the T2I customized model overfits the scene's background, our method often produces a similar overfit (\cref{fig:limitations}).

\section{Conclusions}

Text-to-video models are becoming increasingly powerful and can now generate complex cinematic shots at high resolution. However, the potential use of such models in real-world applications can only be fully realized if the generated content can be incorporated into a larger narrative that contains specific characters, styles, and scenes. Thus, the task of video customization becomes paramount, yet   methodologies that achieve this  are still  underexplored.

In this work, we overcome a major challenge towards this aim, which is the lack of customized video data. We develop a novel framework that  directly translates the immense progress from the image realm to the video realm. Importantly, our method is generic and can be applied to any video model built on a pre-trained T2I model. Our framework unveils the powerful priors learned by T2V models, as evident by the ability to successfully generate motion for specific subjects without ever observing these subjects in motion.

\vspace{-7px}
{ \paragraph{\textbf{Societal Impact}}
Our primary goal in this work is to enable novice users to
generate visual content in a creative and flexible way. However, there is a risk of misuse for creating fake or harmful
content with our technology, and we believe that it is crucial
to develop and apply tools for detecting biases and malicious
use cases in order to ensure a safe and fair use.}

\vspace{-4px}
{ \paragraph{\textbf{Acknowledgements}}
We would like to thank Jess Gallegos, Sarah Rumbley, Irina Blok, Daniel Hall, Parth Parekh, Quinn Perfetto, Andeep Toor, Hartwig Adam, Kai Jiang, David Hendon, JD Velasquez, William T. Freeman and David Salesin for
their collaboration, insightful discussions, feedback and support. We thank owners of images and videos used in our
experiments for sharing their valuable assets (attributions
can be found in our webpage).}

{
    \small
    \bibliographystyle{ieeenat_fullname}
    \bibliography{main}
}

\clearpage
\setcounter{page}{1}
\appendix

\section{Implementation Details}
We demonstrate our method's effectiveness on two prominent inflated models, namely Lumiere~\cite{lumiere} and AnimateDiff~\cite{guo2023animatediff}. For Lumiere, we employ $8$ TPU-v5 cards with a learning rate of $1e-4$ and a batch size of $8$ for training both the Motion Adapters and the Spatial Adapters. The Motion Adapters are trained for $2,000$ steps on $1,600$ videos from Lumiere's training set, while the Spatial Adapters are trained for a maximum of $500$ steps on a combination of customized data and $40$ random videos from the model's training set. Importantly, the Motion Adapters need to be trained for a \emph{minimal number of steps} to remain as close as possible to the frame distribution of the T2V model. We employ the same rank of $4$ for the Motion Adapters and the Spatial Adapters. 

For AnimateDiff, we employ a single A100 GPU (with $40$ GB) with a learning rate of $2e-5$ and a batch size of $1$ for training both the Motion Adapters and the Spatial Adapters. The Motion Adapters are trained for $4,000$ steps on $2,000$ videos from the WebVid~\cite{Bain21web} evaluation dataset, while the Spatial Adapters are trained for a maximum of $600$ steps on a combination of customized images and $40$ random videos from WebVid~\cite{Bain21}.  We employ a rank of $4$ for the Motion Adapters and $64$ for the Spatial Adapters, to adapt our method to the larger channel dimension by AnimateDiff. 

Additionally, we employ a noise offset to accommodate very bright or very dark reference images. For Lumiere, we employ a noise offset scale of $0.01$, while for AnimateDiff, we use a noise offset of $0.1$. The noise offset is duplicated across frames.

\paragraph{ControlNet}
Sec.~4 in the main text showcases the combination of Still-Moving with ControlNet~\cite{zhang2023adding}. This combination allows customizing an existing video according to a personalized subject or customized style, by maintaining its original structure through a depth map or a canny edge map.
In order to obtain a video ControlNet, we inflate a pretrained ControlNet branch by adding Lumiere's temporal layers to it. We then train the spatial ControlNet layers on video data, where the conditions are extracted per frame. This  process results in a video ControlNet branch that can be plugged into Lumiere.

At inference time, 
for generations that require an adaptation of the structure, we apply ControlNet only on the first half of the denoising iterations, and multiply the output of the ControlNet branch by $0.5$ to reduce the structural constraint. 
To obtain the results presented in the paper, we condition the ControlNet branch on either depth maps~\cite{Ranftl2019TowardsRM} or HED edge maps~\cite{Xie_2015_ICCV}. The same condition type is used for all examples of a specific style or subject.

\begin{figure}
    \centering
    \includegraphics[width=0.75\linewidth]{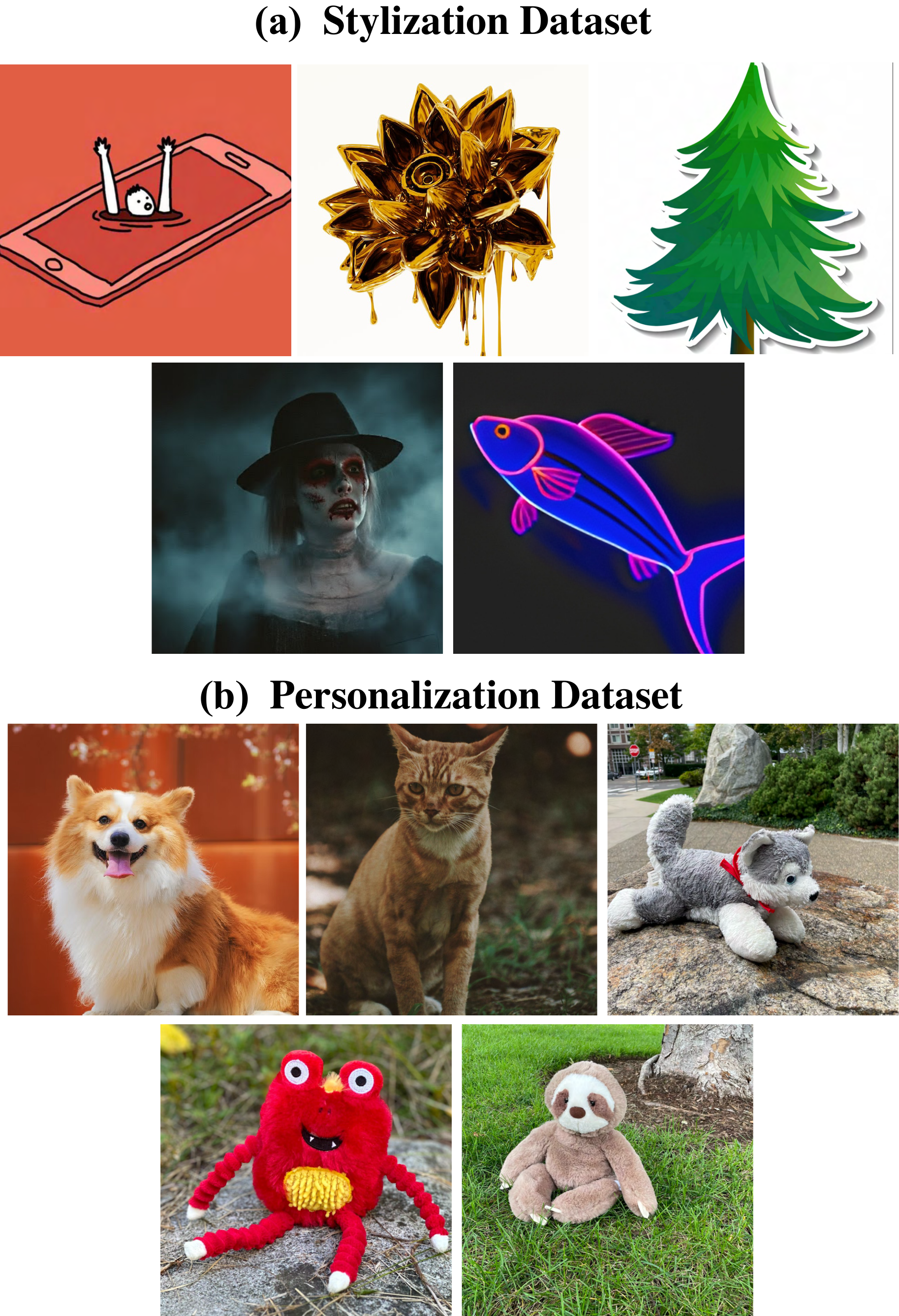}
    \caption{\textbf{Evaluation Dataset.} \textit{We consider $10$ (a) styles and (b) personalized objects. The bottom two images in (a) were generated by Google. The rest are real images from StyleDrop~\cite{sohn2023styledrop} and DreamBooth~\cite{ruiz2022dreambooth}. Our dataset encompasses vector art styles ((a) top) and realistic highly detailed styles ((a) bottom). For the personalization usecase, we consider examples from the Dreambooth dataset~\cite{ruiz2022dreambooth} including realistic characters ((b) top) and OOD challenging characters ((b) bottom).}   }
    \label{fig:dataset}
\end{figure}

\begin{figure*}
    \centering
    \includegraphics[width=0.8\linewidth]{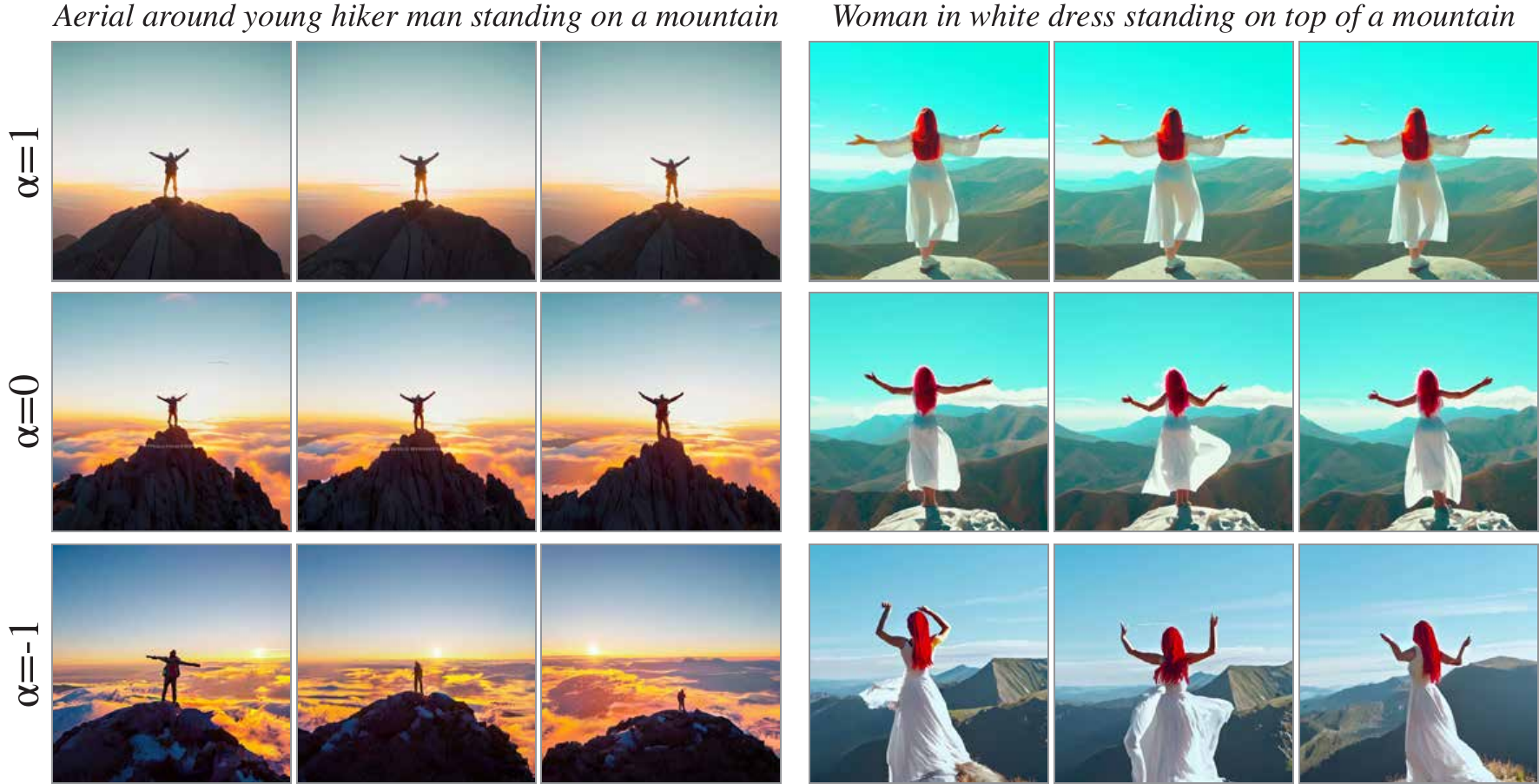}
    \caption{\textbf{Motion Adapters generalization.} \textit{Images above are still frames from videos generated by Lumiere~\cite{lumiere}. We observe that the Motion Adapters generalize their ability to control the amount of motion in the produced videos. When applying $\alpha=1$, the videos freeze, $\alpha=0$ preserves the original behavior of the model, while  negative scales increase the amount of produced motion.} }
    \label{fig:motion_lora}
\end{figure*}

\section{Evaluation Dataset Details}

In Fig.~\ref{fig:dataset} we enclose our full set of reference images for the evaluation of our method and the baselines. We consider $10$ styles and personalized objects. The style reference images are from StyleDrop~\cite{sohn2023styledrop} ((a) top), with $2$ additional realistic styles (bottom two images in (a)). The personalization reference images are from the DreamBooth~\cite{ruiz2022dreambooth} dataset. Our dataset encompasses vector art styles ((a) top) and realistic highly detailed styles ((a) bottom). For the personalization usecase, we consider examples from the Dreambooth dataset~\cite{ruiz2022dreambooth} including realistic characters ((b) top) and OOD challenging characters ((b) bottom).  

\subsection{Evaluation Prompts}\label{app:list_of_prompts}

In the following, we list the prompts used for our automatic evaluation and the user study. The evaluation prompts for the stylization subset are simply extracted directly from Lumiere to facilitate a fair comparison. The personalization prompts were generated by an LLM, with the task of providing diverse prompts containing different locations and actions.

The following are the prompts used for the personalization comparison dataset where [V*] was replaced by the rare token (e.g. \textit{monadikos}) and [c] was replaced with the object class (e.g. dog, cat etc.)
\begin{enumerate}%
    \item \it{"A [V*] [c] driving a race car."}
    \item \it{"A [V*] [c] exploring a flower garden."}
    \item \it{"A [V*] [c] having a tea party."}
    \item \it{"A [V*] [c] flying in the sky."}
    \item \it{"A [V*] [c] in Paris."}
    \item \it{"A [V*] [c] peeking out from inside a blanket fort."}
    \item \it{"A [V*] [c] riding a colorful bicycle through a sunny park."}
    \item \it{"A [V*] [c] walking happily in a pile of autumn leaves."}
    \item \it{"A [V*] [c] wearing a fancy hat, looking adorable."}
    \item \it{"A [V*] [c] wearing sunglasses, looking cool."}
\end{enumerate}

The following are the prompts used for the stylization comparison dataset where [s] replaced with the style name used in StyleDrop (e.g. watercolor painting, sticker etc.)
\begin{enumerate}%
    \item \it{"A bear twirling with delight in [s] style."}
    \item \it{"A butterfly fluttering from flower to flower in [s] style."}
    \item \it{"A dog walking in [s] style."}
    \item \it{"A lion with a majestic mane in [s] style."}
    \item \it{"A dolphin leaping out of the water in [s] style."}
    \item \it{"A penguin dancing in [s] style."}
    \item \it{"A koala munching on eucalyptus leaves in [s] style."}
    \item \it{"A lion with a majestic mane roaring in [s] style."}
    \item \it{"An owl perched on a branch in [s] style."}
    \item \it{"A family of ducks swimming in a pond [s] style."}
\end{enumerate}

\section{Motion Adapters}
The primary goal of the Motion Adapters is to facilitate training on image data. However, we find that they generalize their control over the amount of produced motion to negative scales. Specifically, when applying scales $\alpha<0$, the amount of motion in the generated videos \emph{increases}. This allows users additional controllability over the generated content. Figure~\ref{fig:motion_lora} presents results of applying the Motion Adapters with different scales on Lumiere~\cite{lumiere}- $\alpha=1$ results in frozen videos, $\alpha=0$ preserves the behavior of the original model, and $\alpha=-1$ introduces enhanced motion.

\begin{figure*}
    \centering
    \setlength{\tabcolsep}{0.5pt}
    \addtolength{\belowcaptionskip}{0pt}
    {\small
    \includegraphics[width=0.94\textwidth]{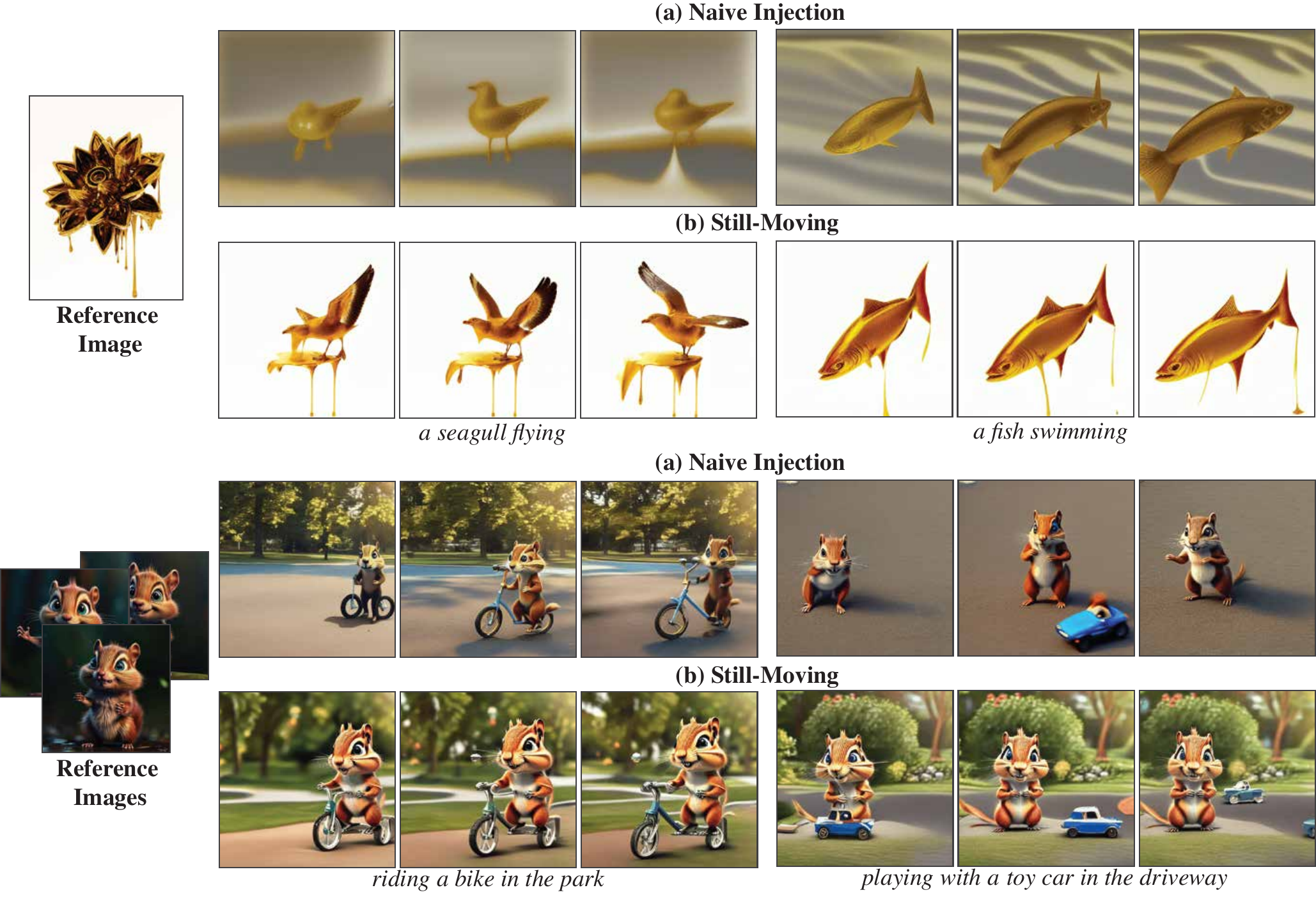}
    
    }
    \vspace{-2px}
    \caption{\textbf{Qualitative results on AnimateDiff.} \textit{We present results of applying Still-Moving to AnimateDiff~\cite{guo2023animatediff} models. We find that for detailed styles and characters, a simple weight injection as performed in~\citet{guo2023animatediff,liew2023magicedit} falls short of adhering to the customized data. Reference images of the chipmunk were generated by Google, whereas the reference image of the flower is real.}}
    \label{fig:animatediff}
\end{figure*}

\end{document}